# Online Model Evaluation in a Large-Scale Computational Advertising Platform

Shahriar Shariat
Turn Inc.
Redwood City, CA
Email: sshariat@turn.com

Burkay Orten
Turn Inc.
Redwood City, CA
Email: borten@turn.com

Ali Dasdan
Turn Inc.
Redwood City, CA
Email: adasdan@turn.com

*Abstract*—Online media provides opportunities for marketers through which they can deliver effective brand messages to a wide range of audiences at scale. Advertising technology platforms enable advertisers to reach their target audience by delivering ad impressions to online users in real time. In order to identify the best marketing message for a user and to purchase impressions at the right price, we rely heavily on bid prediction and optimization models. Even though the bid prediction models are well studied in the literature, the equally important subject of *model evaluation* is usually overlooked or not discussed in detail. Effective and reliable evaluation of an online bidding model is crucial for making faster model improvements as well as for utilizing the marketing budgets more efficiently. In this paper, we present an experimentation framework for bid prediction models where our focus is on the practical aspects of model evaluation. Specifically, we outline the unique challenges we encounter in our platform due to a variety of factors such as heterogeneous goal definitions, varying budget requirements across different campaigns, high seasonality and the auction-based environment for inventory purchasing. Then, we introduce *return on investment* (ROI) as a unified model performance (i.e., success) metric and explain its merits over more traditional metrics such as click-through rate (CTR) or conversion rate (CVR). Most importantly, we discuss commonly used evaluation and metric summarization approaches in detail and propose a more accurate method for online evaluation of new experimental models against the baseline. Our *meta-analysis*-based approach addresses various shortcomings of other methods and yields statistically robust conclusions that allow us to conclude experiments more quickly in a reliable manner. We demonstrate the effectiveness of our evaluation strategy on real campaign data through some experiments.

## I. INTRODUCTION

Advertisers are increasingly exploiting online media to reach target audiences through different channels such as search, display, mobile, video and social advertising. Similar to other forms of marketing, the ultimate goal is to deliver brand messages to the most receptive set of users who are likely to *convert* by taking a desired action after they are subjected to a particular type of ad impression. Even though the definition of a satisfactory outcome is advertiser dependent, identifying a 'likely-to-respond' user in real time is common to all online advertising campaigns. The real-time requirement comes from the fact that a significant chunk of ad impressions (i.e., the opportunity to serve an ad to an online user on a particular publisher page) are sold in marketplaces through auctions held by several ad exchanges [1]. Real-time biding exchanges (RTBs) solicit the availability of an ad impression to all interested parties, such as the advertisers themselves or demand side platforms, like Turn Inc., who manage the advertisers' campaigns on their behalf, and each impression is sold to the highest bidder. Finding the right set of users in this setting requires complex statistical approaches that accurately model user intent and optimize the bid price for every ad impression of each advertiser.

Bid prediction models are at the heart of any advertising technology platform and they need to be enhanced in order to adapt to the evolving requirements of marketers and to remain competitive in a highly dynamic ad marketplace. In other words, new models need to be introduced or the predictive power of the existing models has to be improved continuously. A common first step towards quantifying the efficacy of a new model compared to a baseline is to carry out offline analysis and study the metrics that measure the predictive accuracy such as relative log-likelihood, area under the curve (AUC) and classification accuracy [2] or employ more sophisticated techniques such as contextual multi-armed bandits [3]. Even though offline analysis provides helpful guidance, it does not capture the true performance of a bid prediction model, which needs to operate in a dynamic auction environment where the cost of impressions plays a significant role. No matter how positive the outcomes of the offline experiments are, we need to evaluate the real (i.e., online) performance of a new model on live traffic and on real users in a controlled manner. Online evaluation of a model in our platform is not a trivial task and we need to ensure that our experimentation and evaluation framework is able to:

- handle **traffic allocation** to both the baseline and experimental models,
- allow **dynamical control** of traffic allocation such that the models do not influence each other,
- clearly identify a **performance evaluation metric** which aligns with the advertiser performance expectations as well as business value,
- and, most importantly, enable **model evaluation** in a complex, dynamic and heterogeneous auction environment such that we can make quick progress towards deciding whether to replace the baseline model with the experimental one or not.

Since an algorithmic change in a model affects the behavior of the system we need to consider its impact on every advertiser and make sure that we statistically improve the performance of a quite significant portion of our clients.

Other works in the area of online advertising focus on the experimental design of single test [4], [5]. However, our problem is to design a framework that can reliably and robustly summarize the effect of an improvement in our bid prediction models. We need methods and apparatus that can help us study the effect of a treatment applied to several experiments and summarize them in a principled manner. In statistics literature, meta-analysis provides the basic tools and techniques for such task [6], [7], [8].

In this paper, we present a framework that we utilize for online evaluation of bid prediction models in our platform. We outline the important components of our framework, that are critical for conducting multiple model experiments at a very large scale such as user-based traffic allocation, dynamic traffic control and budget management across different models. We focus on the definition of a performance metric and the model evaluation criteria. More specifically, we identify *return on investment* (ROI) as the most suitable evaluation metric in our setting due to its alignment with both performance and business expectations. Then, we describe our *meta-analysis-based model evaluation* approach, which is robust to outliers and allows us to make more consistent decisions about the online efficacy of a model.

The paper is organized as follows. We first, in § II, describe the necessary background on the ad selection problem and the complications of a large-scale advertising platform. Then in § III, we detail the structure of our experimentation platform and its necessary elements. § V presents our evaluation methodology and § VI studies a practical experiment. We finally conclude this paper by § VII.

## II. BACKGROUND

A demand side platform, manages marketing campaigns of several advertisers simultaneously by identifying the target audience matching the campaign requirements and making real-time bidding decisions on behalf of the advertisers to purchase ad impressions at the 'best' price. In this section, we briefly describe relevant background information on some important aspects of online campaign management in our platform, which will set the stage for our model evaluation framework.

### A. Bid Prediction in Auction Environment

Majority of the ad impressions in the online setting are sold through RTBs as we briefly explained in § I. These real-time decisions are made in response to what are called ad call requests from RTBs. Ad requests come to our system (essentially) in the form of $(user, page)$. Here we use *page* to represent the media on which the ad will be served (e.g., a webpage, mobile app or a video player). The main purpose of ad selection is to identify the most relevant match among all qualifying ads. Let $\mathcal{A} = \{a_1, a_2, ..., a_m\}$ denote the set of all active ads in our system. Identifying the best ad entails finding the ad with the highest bid price for the incoming request:

$$ad^* = \arg\max_{k=1,...,m} Bid(a_k) \qquad (1)$$

Bid price of each ad is a function of campaign parameters and the desired outcome that each marketer is trying to accomplish. Some campaigns are tailored for pure branding purposes rather than expecting a direct response from the user and in this scenario, advertiser usually resort to a fixed bid price. In other words, bids are not *dynamically adjusted* for each request. However, the more common scenario is to have a performance goal associated with the campaign and *optimize* the bid price at each request to maximize the likelihood of getting a desired 'outcome' once the ad impressions are served to online users. Bid estimation in this case involves utilizing machine learned models [9], [10] to estimate the likelihood of the desired event and bid for $a_k$ can be computed as:

$$Bid(a_k) = Goal(a_k) * p(outcome|user, page, a_k) \qquad (2)$$

where $Goal(a_k)$ is a target value that is assigned by the advertiser at the beginning of a campaign and *outcome* is one of a click, a conversion, a view, etc. Intuitively, this computation captures the *expected value* of an impression to a particular advertiser. Collectively, we refer to the system that is capable of 1) estimating likelihoods, 2) computing bids for various types of ads and 3) ranking them according to business and performance objectives as the *bid prediction model*. After we select the best ad internally, we submit a response to the exchange where a secondary auction is held among all participants to identify the winning ad to be served to the online user. We should emphasize that RTB auction is held independently and our internal bid prediction models are required to handle the price uncertainty associated with a very dynamic external marketplace.

### B. Various Campaign Goals

As we briefly touched upon in § II-A, the overall bid prediction system and bidding models are highly dependent on the outcome event advertisers are interested in and the goal value they associate with that event. This outcome could simply be the user clicking on the served ad or it could be a more direct response such as a conversion event (e.g., signing up for a newsletter, requesting a quote or purchasing an item) as identified by the marketer. In other words, we need to keep track of multiple event types (clicks, conversions, engagement, viewability of the ad etc.) and their respective value for each advertiser. While this custom setup gives a lot of flexibility to accomplish various marketing objectives, it results in a highly heterogeneous environment for the optimization platform to evaluate. Commonly used metrics such as *click-through rates* (CTR), *conversion rates* (CVR), cost-per-click (CPC) or cost-per-conversion (CPCV) do not fully capture the global performance of our bidding models. For example, as we try to improve the CTR-based performance, we sometimes find ourselves hurting the CVR metric for some campaigns or the average cost of conversions may go up. Model experimentation in this environment requires a unified (i.e., normalized across campaigns) goal definition that we can rely upon to quantify and assess the overall model performance.

### C. Seasonal and Varying Budgets

In addition to the target audience a campaign is trying to reach, two other factors that play a crucial role in a campaign's final outcome are the duration of the campaign (i.e., flight time) and the total allocated budget. Depending on the time of the year, availability of a product to be promoted or their business strategy, marketers modify campaign goals and budgets continuously. At any given time, we have in our platform several thousand campaigns active that have competing goals as well as highly varying budgets. Disparity between the budgets of a small campaign and a larger one can easily reach a factor of 1000s. Since our bidding models are simultaneously optimizing the campaign performance over the whole ecosystem, composition of different types of campaigns we are running, their goals and respective budgets play a fundamental role in our model evaluation. In other words, our evaluation approach has to take into account all these factors while making a final decision on the outcome of an online experiments.

| Model A: campaign level statistics |||| 
| Campaigns | Value Generated | Overall Spending | Return on Investment (ROI) |
|---|---|---|---|
| $C_1$ | $Value_{A,1}$ | $Spend_{A,1}$ | $ROI_{A,1} = Value_{A,1}/Spend_{A,1}$ |
| $C_2$ | $Value_{A,2}$ | $Spend_{A,2}$ | $ROI_{A,2} = Value_{A,2}/Spend_{A,2}$ |
| $C_3$ | $Value_{A,3}$ | $Spend_{A,3}$ | $ROI_{A,3} = Value_{A,3}/Spend_{A,3}$ |
| ... | ... | ... | ... |
| $C_n$ | $Value_{A,n}$ | $Spend_{A,n}$ | $ROI_{A,n} = Value_{A,n}/Spend_{A,n}$ |

| Model B: campaign level statistics |||| 
| Campaigns | Value Generated | Overall Spending | Return on Investment (ROI) |
|---|---|---|---|
| $C_1$ | $Value_{B,1}$ | $Spend_{B,1}$ | $ROI_{B,1} = Value_{B,1}/Spend_{B,1}$ |
| $C_2$ | $Value_{B,2}$ | $Spend_{B,2}$ | $ROI_{B,2} = Value_{B,2}/Spend_{B,2}$ |
| $C_3$ | $Value_{B,3}$ | $Spend_{B,3}$ | $ROI_{B,3} = Value_{B,3}/Spend_{B,3}$ |
| ... | ... | ... | ... |
| $C_n$ | $Value_{B,n}$ | $Spend_{B,n}$ | $ROI_{B,n} = Value_{B,n}/Spend_{B,n}$ |

Fig. 1. Important campaign statistics used in model A/B experiment. Both models $A$ and $B$ deliver impressions for $n$ campaigns and the resulting ROI statistics are combined to make an overall model assessment.

## III. EXPERIMENT PLATFORM SETUP

Offline experimentation and evaluation using classification or ranking metrics is a commonly used initial step towards improving a bidding model. However, we have experienced multiple times in various experiments that offline success does not necessarily indicate positive outcomes in live traffic and finding correlations between offline and online results is an ongoing area of research (cf. [11]). This is particularly, due to lack of access to the exchange's logs and therefore, the inability to accurately simulate the external auction. Furthermore, for practical reasons, the logs typically cannot contain the data for all bid-requests and only the portion that are won by the DSP is kept.

The most commonly used approach for online evaluation is A/B testing, which is a randomized experiment where a new *treatment* model, referred to as $B$, is being evaluated against a *control* (or baseline) model, referred to as $A$. The goal is to compare the efficacy of a treatment model $B$ over the control $A$ on a **statistically identical population** by measuring the difference in an **overall evaluation criteria** of interest on (dynamically adjustable) **online user traffic**. Running controlled online experiments at internet scale is an interesting problem by itself (cf. [4], [5]) but a detailed description of such an experimentation platform is beyond the scope of our paper. Instead our focus in this section is to highlight the most important aspects of our online experimentation platform as it pertains to our specific needs of evaluation for bidding models.

### A. Online Traffic Allocation

The success of a randomized experiment heavily relies on a proper split of the samples over which the control and treatment models will be evaluated. In our context, one may consider allocating the incoming live (request) traffic randomly to one of the bidding models and let the chosen model handle the selection of the best ad from one of the active campaigns as well as the proper choice of a bid price. However, this randomization scheme would ignore the user component of a bid request completely and would result in both the control and treatment model serving impressions to the same user. Subsequently, it would be difficult to identify the actual causal impact of a model in case of a successful outcome such as a conversion. In order to minimize the impact of models over each other, a *user-based traffic split* is best suited for our A/B testing purposes [12]. More explicitly this corresponds to pre-allocating a subset of users to the treatment (i.e., $B$) model such that any time we receive a bid request from *any* RTB for this user, ad selection and bid prediction will always be handled by model $B$ and the rest of the requests will be directed to the baseline model $A$.

After deciding on how to allocate the incoming online traffic to models, the next decision to make is *how much traffic* to actually allocate to the treatment model. In an ideal A/B testing scenario, user sample would be split equally at $A : \%50, B : \%50$. Since the live experiments are conducted in a production environment while the campaigns are active, system health and campaign performance requirements can potentially be jeopardized if a model exhibits unexpected behavior with high traffic. It is much safer to activate the treatment model on a small percentage of users initially and ramp up the traffic as we collect evidence indicating that the experimental model is performing better then the control. For example, one can decide to adopt the following schedule while running a model experiment:

Online Experiment Schedule Phase 1
  Traffic allocation: $A : \%99, B : \%1$, duration: 1-2 days
Online Experiment Schedule Phase 2
  Traffic allocation: $A : \%90, B : \%10$, duration: 1 week
Online Experiment Schedule Phase 3
  Traffic allocation: $A : \%80, B : \%20$, duration: 1 week
Online Experiment Schedule Phase 4
  Traffic allocation: $A : \%50, B : \%50$, duration: 1 week

Actual traffic percentages and the duration of each phase should be selected such that *sufficient evidence* is collected over all representative entities in the experiment. In our platform, even though our ultimate goal is to make a model-level assessment, we consider every active campaign as a separate experiment under the new model (see Fig. 1). Therefore, we need to tune the traffic share and the duration of the experiment such that the majority of the campaigns can collect sufficient number of users to exhibit enough statistical power [12]. We should emphasize that treating each campaign as an individual experiment is a very fundamental aspect of how we approach online model evaluation. Recall from § II that every campaign object is identified by a different target audience, flight time, optimization goal and a monetary budget. Given this variability, treating various types of active campaigns while running experiments requires a meticulous evaluation methodology so that we are able to make statistically reliable and robust conclusions. We should also point out that the campaign level assessment approach we will outline does not change how we split and allocate the online traffic based on user IDs. Since the rate at which we receive queries from multiple RTBs in our platform is very high (~2.5 million bid requests per second as of this writing), with a high enough traffic percentage and a reasonably long experiment (e.g., 2-3 weeks) majority of our active campaigns get to observe sufficient number of requests from many users. In other words, uniform random split criteria yields a good (i.e., representative) sample for both models $A$ and $B$ under each campaign such that not only can we make individual assessment at the campaign level, but we can also make a statistically reliable decision for the overall performance of a bidding model.

### B. Overall Evaluation Criteria

Another important component in an online experiment is the pre-determined *overall evaluation criteria* (OEC), which will be used to determine evaluation outcome. Even though it is possible to track multiple metrics during a large scale experiment to gain insight into the performance of a model from different perspectives, OEC should be identified as the ultimate metric that both captures the statistical properties of our models and also aligns well with the business success. As we discussed in § II-B, flexibility in defining custom campaign goals makes it a challenge for us to compare the impact of a new bidding model across thousands of active campaigns. As an example, our bidding algorithm might be positively impacting (i.e., reducing) the cost of conversion for a conversion optimizing campaign whereas a click optimization type campaign might be observing an inferior CTR within the same time period. Our aim is to define a unified metric that will simultaneously capture the performance of every type of campaign which is adaptive to their custom defined goal types.

Before we define our choice of OEC metric, we first have to introduce the concept of a *value* that each campaign may get to observe after each ad impression is served. Recall that campaigns can be tracking various types of events such as clicks, conversions or ad views. Advertisers define a goal value that they associate with the event they are optimizing towards. Based on the event type and the monetary value they associate with the ad, we can define whether they generated value as a result of an impression or not. For example, if the target event is a click, which has been valued at $0.5, and if the campaign generates 1000 clicks after a period of time, we can conclude that $500 worth of value was generated within that campaign period. Then, for campaign $i$ which is tracking $e$ different types of events valued respectively at $value_1, value_2, ..., value_e$ overall value generated after spending a budget of $spend_i$ can be computed as:

$$Value_i = \sum_{k=1,...,e} value_k * (\# \text{ of events of type } k) \quad (3)$$

In other words, we can say that campaign $i$ generates a total $Value_i$ as a result of investing $Spend_i$ in ad impressions. Then we can define the overall success level of a campaign in terms of *return on investment* (ROI):

$$ROI_i = \frac{Value_i}{Spend_i} \quad (4)$$

Unlike the other event specific metrics such as CTR, CVR, CPCV, etc., ROI provides a normalized performance metric that unifies the overall success level of any type of campaign regardless of their target event. ROI is our choice of OEC for our online bidding model experiments and we will demonstrate in the section below how it can be used for A/B test evaluations.

### C. Evaluation Setup Definitions & Notation

Having discussed the traffic split and OEC in detail in previous section, we would like to outline all important quantities we measure at the campaign level that will be used for overall model evaluation. Note that we use online experiment and A/B test interchangeably and they both refer to the randomized experiment we conduct in a production environment on live bid request traffic. Listed below is our notation and the definition of relevant quantities we will be using in § IV (also see Fig. 1).

- i=1,...,n: index that identifies different campaigns which are active during an online experiment
- $A$: baseline model currently used in production
- $B$: experimental model we are evaluating against A
- $M \in \{A, B\}$ : indicates a model, A or B
- $Spend_{M,i}$ : Total spending of campaign $i$ by model $M$
- $Value_{M,i}$ : Total value generated for campaign $i$ by model $M$ after spending $Spend_{M,i}$ during the experiment
- $ROI_{M,i} = \frac{Spend_{M,i}}{Spend_{M,i}}$: ROI of campaign $i$ observed by model $M$

The overall evaluation pipeline is illustrated in Fig. 2. Note that the traffic increase to a model might need the approval of a manager who has to weigh the estimated improvement caused by the model against its computation cost and other considerations. Particularly, for this purpose we need an evaluation method that gives a reliable *improvement amount*. Before we conclude this section, we should reemphasize that campaign level ROI is a core element of our evaluation method where each campaign's spend and ROI statistics under each model are treated as an individual experiment and the final outcome is based on a robust combination of individual ROIs. The next two sections we mention different evaluation designs that we have considered and describe our proposed and final approach in detail.

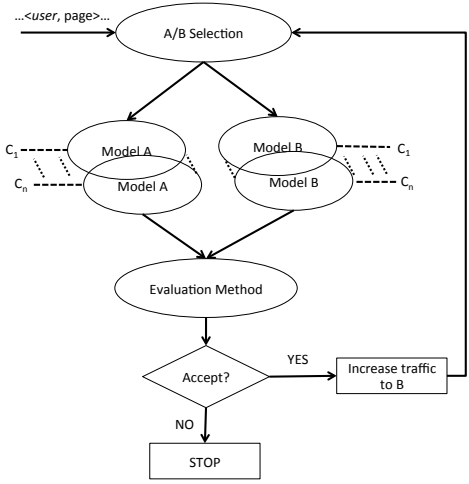

Fig. 2. The overall pipeline of the evaluation platform. We increase the traffic to the treatment model until it becomes dominant and replaces the baseline. The traffic increase might be subjected to the approval of a manager.

## IV. EXISTING EVALUATION DESIGNS

The problem is to design an evaluation method that can quickly and reliably reject or accept a new model for assigning more traffic. Moreover, we would like to gain as much insight as possible into the performance of the proposed models. Furthermore, the evaluation method must be able to estimate the expected amount of improvement.

**Micro-averaging (Micro)**: This is a conventional summarization technique in data mining and information retrieval [13]. Based on this method, We define the ROI of the model as

$$ROI_M = \frac{\sum_{i=1}^{n} Value_{M,i}}{\sum_{i=1}^{n} Spend_{M,i}} \quad (5)$$

Note that (5) can be written as

$$ROI_M = \sum_{i=1}^{n} \frac{Spend_{M,i}}{\sum_{i=1}^{n} Spend_{M,i}} \frac{Value_{M,i}}{Spend_{M,i}}, \quad (6)$$

which is the weighted average of the ROI of each campaign (for each model) by its spend. The final decision using this method is made based on the difference of two computed ROIs. That is,

$$\mu^{Micro} = ROI_B - ROI_A \quad (7)$$

and we require $\mu^{Micro} > \theta$, where $\theta$ can be set using an A/A test [12].

*A/A test*: To decide on $\theta$, we randomly divide the data (bid requests) of each campaign observed by the control model (A) into two non-overlapping subsets, $A_1$ and $A_2$, proportional to the size of A and B and compute $\mu_{A/A}^{Micro} = ROI_{A_1} - ROI_{A_2}$ as if the subsets were two models. Since the two subsets are created from the same model, a non-zero $\mu_{A/A}^{Micro}$ can be ascribed to the system variation. We repeat this task K times (5 in our experiments) and we set $\theta = \frac{1}{K}\sum_{k=1}^{K} \mu_{(A/A)_k}^{Micro}$. Note that the A/A test accounts for the variation of the whole system rather than for each individual campaign.

First, it must be noted that our objective is to improve the performance of as many campaigns as possible while obtaining a better global picture. In other words, if we observe a global advantage such that a few large campaigns experience very large improvement in their metrics while many other campaigns suffer from a degradation, then we must conclude that the model is not significantly better, compared to the control. This method naturally gives more weight to the higher spending campaigns. In § VI we will demonstrate how this can be a dangerous property.

**Macro-averaging (Macro)**: Another conventional data mining method closely related to Micro-averaging is Macro-averaging. It is essentially the average of the differences of metric over campaigns. That is,

$$\mu^{Macro} = \frac{1}{n}\sum_{i=1}^{n}(ROI_{B,i} - ROI_{A,i}). \quad (8)$$

The final decision using this method is that we accept the model if $\mu^{Macro} > \theta$, where $\theta$ is determined using an A/A test analogous to Micro-averaging.

This method has the advantage of considering the performance of the treatment model equally over all campaigns. The obvious shortcoming of this method is its sensitivity to the outliers. A large difference between the performance of the two models can happen due to the traffic split especially for small campaigns. This can result in an undesirable skew in the measure. Another disadvantage of this method is considering any difference without accounting for the variation and sampling error within and between the campaigns.

A plausible solution for reducing the sensitivity to outliers is to use median instead of average. However, since we usually have many campaigns, the median and average are not far different from each other. In our experiments we rarely noticed a difference between median and average such that it results in contrasting decisions. Other outlier removal and noise reduction techniques are also not suitable since removal of any of the campaigns from the analysis might result in a biased decision making. One needs to weigh different measurements based on their reliability and then combine them.

**Summarized statistical analysis**: There are two possible methods for summarizing the statistics of the effect of the treatment model: *i)* One can compare the distributions of the $ROI_{A,i}$ and $ROI_{B,i}$ over all campaigns using a statistical test such as Kolmogorov-Smirnov (K-S) or other fitting tests [14]. Such tests, again, ignore the variation inside each experiment and also do not result in an overall improvement measure. *ii)* To account for the variation within each campaign one can use a statistical significance analysis such as a Student-t test [14] by considering the distribution of the measurements within each campaign. This method however, still leaves us without a wholesome measure of improvement. Also, since we conduct the tests on a pretty large set of campaigns, we need to deal with the family-wise error and false discovery rate issues[15], [16]. Another disadvantage is that this approach leaves the insignificant campaigns out of the analysis while this is not desirable. We need a method that systematically accounts for the variations and combines the statistics of all campaigns according to that. Meta-analysis provides us with such tools. We describe the detail of our adaptation of meta-analysis in the next section.

## V. Proposed Meta-analysis

Meta-analysis enables us to combine our estimates of the difference of the control and treatment models over multiple campaigns. In this section, we describe the methodology that provides us with a global estimate based on which we decide whether to reject the model or increase its traffic share. We divide the traffic that is received by each model into several parts based on the user ids or uniformly at random. Each part, for instance, can correspond to 1% of the traffic directed to the model. To this end, assume that models $A$ and $B$ receive $m_A$ and $m_B$ parts of the traffic, respectively. Also, to simplify the notation we denote the measurement (ROI in this case) of the $j$th part of the $i$th campaign by $R_{A,i,j}$ and $R_{B,i,j}$ for models $A$ and $B$, respectively. The parts might needed to be excluded from the analysis due to insufficient number of impressions as a pre-processing procedure. One such procedure is described in § VI.

### A. Effect size

The effect size of a test is a (standardized) measurement of the difference between the means of the two sets of samples [6]. This gives us an estimate of how the distributions of the two sets (assuming a normal distribution) differ. Assuming that one set of samples belong to our treatment and the other is the result of applying the control, the effect size can estimate how much better or worse a treatment model is compared to the control. Combining these effect sizes can provide us with a global picture of the performance of the treatment.

In statistical modeling three major types of models are considered. Fixed, random and mixed effect models. Fixed effect model assumes that the effects are the same in all tests and the difference between them is only caused by sampling error. In contrast, the random model assumes that the variation is systematic and the effects are sampled from a distribution related to the structure of the tests. A mixed model, obviously, assumes a mixture of the two. We assume a random effect model because the variation between campaigns can be explained by assuming a latent distribution governing the local and global auctions and the inherent structure of the campaigns themselves. We start with the fixed effect model to calculate the effects and their corresponding variances and then expand to the random effect model by computing the between-variance of the effect sizes.

Let $\bar{R}_{M,i}$ and $s^2_{M,i}$ be the sample average and variance of $R_{M,i,\cdot}$ over all acceptable parts of the model $M$ for campaign $i$. The effect size is the normalized difference of the means of the two distributions. Therefore, assuming a normal distribution over the measures, ROI of each part in our case, the standard effect size of the $i$th experiment [1] is defined as $\delta_i = \frac{\bar{R}_{B,i} - \bar{R}_{A,i}}{s_{p_i}}$, where $s^2_{p_i} = \frac{(m_A-1)s^2_A + (m_B-1)s^2_B}{m_A + m_B - 2}$ is the pooled variance of $R_{A,i,\cdot}$ and $R_{B,i,\cdot}$ distributions. The pooled variance is the weighted average of the sample variances of the two populations such that the weights are their sizes. The assumption is that both populations share the same variance but their means might be different [8]. Note that $\delta_i$ follows a noncentral Student-t distribution with $df = m_A + m_B - 2$ degrees of freedom and noncentrality parameter, $\delta$ [17]. Therefore, for each campaign, the unbiased effect size (mean of the distribution) is $d_i = c(df)\delta_i$, where $c(df) = 1 - \frac{3}{4df-1}$. The corresponding variance of the effects is also the variance of the noncentral Student-t distribution $v_i = c(df)^2(\frac{m_A+m_B}{m_A m_B} + \frac{d_i^2}{m_A+m_B})$. For a comprehensive discussion on the effect sizes refer to [18], [6].

---
[1] We will use experiment or study and campaign interchangeably.

Note that the variance is at its minimum when $m_A = m_B$. Also, note that the variance is quadratically related to $d$. Therefore, when the allocated traffic to the treatment model is much smaller than the control, the variance is high. Similarly, when the effect itself is large, our uncertainty about its accuracy is high.

The next step is to combine the effects. We approximate the noncentral Student-t distribution of the effects by Gaussian. Furthermore, it is reasonable to assume that the experiments are mutually independent since the campaigns do not share ads or budget. The total effect can be summarized by the weighted average of the effects of each individual experiment. That is,

$$T = \sum_{i=1}^{n} \beta_i U_i, \quad (9)$$

where $U_i \sim \mathcal{N}(d_i, \sqrt{v_i})$ and $T \sim \mathcal{N}(\mu, \sqrt{\nu})$. Note that $\mu = \sum_{i=1}^{n} \beta_i d_i$ and $\nu = \sum_{i=1}^{n} \beta_i^2 v_i$. It is straightforward to observe that setting $\beta_i = \frac{1/v_i}{\sum_{i=1}^{n} 1/v_i}$ minimizes the variance (uncertainty) of $T$, subject to $\beta_i > 0$ and $\sum_{i=1}^{n} \beta_i = 1$ [18]. Therefore, letting $w_i = \frac{1}{v_i}$, one could show that the total mean effect size and variance are

$$\mu = \frac{\sum_{i=1}^{n} w_i d_i}{\sum_{i=1}^{n} w_i} \text{ and} \quad (10)$$

$$\nu = \frac{1}{\sum_{i=1}^{n} w_i}, \quad (11)$$

respectively. Equation (10) is intuitively an appropriate estimator of the summary effect for us since it accounts for the variation inside each experiment and weighs them based on our uncertainty about their effect sizes. Furthermore, (11) accounts even for the traffic directed to the treatment such that the total variance is higher for more unbalanced traffic allocations between treatment and control.

Now, we need to study the possible variation between the studies and test our assumption of a random model. Namely, we need to test whether the variation of the effects are due to sampling or there might be an underlying distribution causing such variation. The homogeneity statistics, Cochran's $Q$ [19], is the appropriate and conventional homogeneity test, which addresses this question. The null hypothesis is that all effect sizes are the same and the observed variation is only due to sampling error. Cochran's $Q$,

$$Q = \sum_{i=1}^{i=n} w_i (d_i - \mu)^2, \quad (12)$$

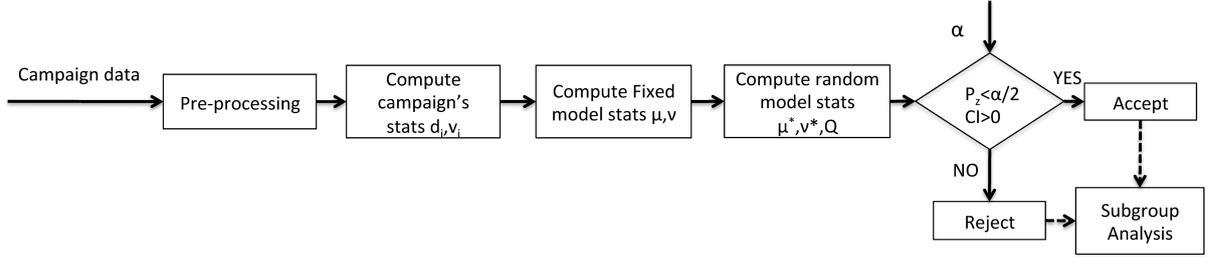

Fig. 3. The steps for making a decision by our method. Subgroup analysis might be omitted where a 'strong' rejection is suggested by the method.

is distributed as $\chi^2$ with $n-1$ degrees of freedom. Therefore, $p_Q = 1 - \chi^2(Q, n-1)$ defines the p-value of the homogeneity. This is an important metric that shows whether we are facing a homogeneous set of effects or not. A homogeneous effect is rather desirable since it stands for a steady and explainable effect over all tests. It means that the effects have enough overlap that we can safely say that the difference between them looks like to be the result of sampling and the model affects all the tests roughly in the same way.

We now compute the statistics of the random effect model. Herein, we assume that the effect sizes are sampled from a normal distribution $T^{between} \sim \mathcal{N}(0, \tau)$. The variance of this distribution is defined by

$$\tau^2 = \begin{cases} 0 & Q < (n-1) \\ \frac{Q-(n-1)}{\lambda} & otherwise \end{cases} \quad (13)$$

where $Q$ is given by (12) and $\lambda = \sum_{i=1}^{n} w_i - \frac{\sum_{i=1}^{n} w_i^2}{\sum_{i=1}^{n} w_i}$. The above is based on the simple method of moments [7], which is the most popular method to compute $\tau^2$. Based on the random model, we are assuming that $U_i^* = U_i + T^{between}$, where $U_i^* \sim \mathcal{N}(d_i^*, v_i^*)$, where $d_i^* = d_i$ and $v_i^* = v_i + \tau^2$. Consequently, $w_i^* = \frac{1}{v_i^*}$. Therefore the total effect is $T^* \sim \mathcal{N}(\mu^*, \nu^*)$ where

$$\mu^* = \frac{\sum_{i=1}^{n} w_i^* d_i}{\sum_{i=1}^{n} w_i^*} \text{ and} \quad (14)$$

$$\nu^* = \frac{1}{\sum_{i=1}^{n} w_i^*}. \quad (15)$$

The above approach is rather robust to outliers since it assigns a smaller weight to those campaigns that report abruptly high or low effect sizes. Furthermore, the variance of the estimation can grow if the effects are widely spread. These are exactly the properties that we were seeking for our estimation of the overall effect of a treatment model. That is, the outliers are systematically downgraded based on their uncertainty and the variation between the effects is also taken into account.

The null hypothesis that the summary effect is zero can be tested using a Z-value such that

$$Z = \frac{\mu^*}{\sqrt{\nu^*}}. \quad (16)$$

Particularly, the effect is significant if $(P_z = (1 - \phi(|Z|))) < \frac{\alpha}{2}$, where $\phi^{-1}(\cdot)$ is the inverse cumulative distribution function of the normal distribution. The confidence interval [14] for the confidence level $\alpha$ is defined as

$$CI = [\mu^* - \phi^{-1}(1 - \frac{\alpha}{2})\sqrt{\nu^*}, \mu^* + \phi^{-1}(1 - \frac{\alpha}{2})\sqrt{\nu^*}]. \quad (17)$$

The primary decision for accepting or rejecting the treatment model is based on the above significance test and the confidence interval. Based on the purpose of the model and the allocated traffic, we require a certain level of significance for the overall effect to be accepted. The significance level can be adjusted using an A/A test as described in § IV. Furthermore, the confidence interval gives an estimation of the expected improvement. Usually, we require a smaller significance level when the traffic share of the treatment model is smaller.

The sequence of the decision-making for the proposed evaluation method is illustrated in Fig. 3

### B. Subgroup analysis

It is necessary, in many cases, to explain the effect of the model on a subset of campaigns and then study the relationship between them. For instance, we might need to know how the model affects the high spending campaigns or those that seek a conversion or click goal. Furthermore, one is usually interested in understanding the variation inside and between subsets or groups. Particularly, we are interested in observing whether or not the group membership can explain the variation among the effects. The analysis of the variation amounts to analysis of Cochran's $Q$ for different groups, which is typically referred to as *subgroup analysis* in the literature [6].

Let $G = \{g_1, \cdots, g_K\}$ be the set of subgroups such that $\bigcup_{k=1}^{K} g_k = \{C_1, \cdots, C_n\}$ and $g_k \cap g_{k'} = \varnothing \quad \forall k, k' \in \{1, \cdots, K\}, k \neq k'$. One then, can compute $Q_k^*$ using a formulation similar to (12) modified so to use the summary effect and weights of the random model. Essentially, we want to study the homogeneity of the effects around the mean within each subgroup. That is,

$$Q_k^* = \sum_{C_l \in g_k} w_j^*(d_l - \mu_k^*) \quad (18)$$

where $\mu_k^*$ is the summary effect of the random model for the $k$th subgroup. The total $Q^*$ is also computed over all studies given the total mean effect, which in turn represents the deviation from the grand mean. Now, we would like to test whether the total heterogeneity of the studies can be explained by aggregating the heterogeneity of all subgroups. Remembering that $Q$ follows a $\chi^2$ distribution, we define

$$Q_{within}^* = \sum_{g_k \in G} Q_k^*. \quad (19)$$

Finally, if the summation of the homogeneity statistics was able to explain the total homogeneity, one might conclude that there is no considerable variation between the groups and thus the group membership does not define the behavior of a given campaign. To represent that, the variation between the subgroups can be calculated as

$$Q^*_{between} = Q^* - Q^*_{within} \tag{20}$$

and tested for significance with the null hypothesis that the group membership does not affect the effect sizes. $Q_{between}$ has $K-1$ degrees of freedom. The same study can be done in a hierarchical fashion to study larger subgroups [6].

Using the above approach, it is possible that we find model $B$ to be effective only in certain cases. For instance, the campaigns that have a very low budget might benefit from the properties of the proposed model. In that case, we might decide to add $B$ to our stock of models and apply per business requirements.

## VI. CASE STUDY

The purpose of the this study is to show that our proposed method [2] is able to arrive at a robust and consistent decision using a smaller portion of the traffic compared to other methods. It is intuitive that a more reliable comparison can be achieved when the treatment model receives more traffic. Therefore, a good measurement mechanism is the one that arrives at a decision that does not change as we increase the traffic directed to the treatment model. Essentially, we demonstrate that our meta-analysis-based approach points to a decision in the first step of the analysis while other methods need more traffic (samples) to arrive at the same decision.

In this case study, we are studying the effect of a change that affects a primary part of our prediction algorithm. The model will be applied to the campaigns that intend to serve ads on mobile devices. The hope is to observe a significant increase in the ROI of the whole system. To comply with company policies we refrain from sharing the absolute numbers and restrict ourselves in reporting only relative results. The number of campaigns that are studied is about a thousand.

We use a noise removal procedure before applying our evaluation method. We require each part to contain a minimum number of impressions (100 in this experiment) and if it does not, then we remove that specific part from the analysis. If the number of removed parts for each model exceeds 10% of the total parts of each model, then the whole campaign is disqualified. Essentially, the number of qualified parts for model A and B must be more than $0.9m_A$ and $0.9m_B$, respectively so that the campaign is included in the analysis. This means that tests might have different sizes depending on how much budget they have and how many users they can reach. We observed that only 0.5% of the campaigns were disqualified as the result of this pre-processing. To be fair to other methods we apply them to the same set of campaigns.

**Setting 1:** Two experiments are conducted to observe the robustness and correctness of the decision-making procedure by each method. The experiments differ in the traffic share that the treatment model receives and in the time period in which they were running. In particular, in the first experiment the treatment model receives 10% while control receives the rest of the traffic and it runs for 12 days. The second experiment runs right after the first experiment for 10 more days but the treatment model receives 20% of the traffic. In this setting, we constrain ourselves to the common campaigns in both experiments and make sure that their setup (budget, targeting, etc.) is not changed between the two time periods. In other words, the campaigns that are under study will be limited to the intersection of the two sets of campaigns that were present in both experimentation periods and also pruned for identical setup. This constraint removes nearly 32% of campaigns from the union of the two sets of campaigns. Most (98.4%) of the removed campaigns are the ones that did not exist in the first experiment. The rest of the removed campaigns, mainly, had budget changes. The whole experiment runs over a period of one month to make sure the seasonality does not affect our results substantially. This approach will allow us to observe the robustness of the decision that is made by each method given the increased traffic. The parameters of each method is estimated based on an A/A test as described in § IV. There was almost no change in the computed parameters between 10% and 20% traffic.

[2]The code and a sample data set is available at https://github.com/turn/ModelEvaluation

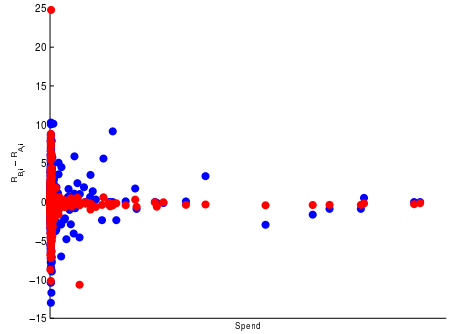

Fig. 4. The difference of ROIs of the two models vs the spend under 10% (blue) and 20% (red) traffic for every *common* campaign between two experiments. The spend is the total spend of the campaign summing up over both models. A few high spending campaigns with slightly worse situation causes the Macro method to change its decision.

The results are depicted in Tab. I. Table I reports the statistics of each method given the parameters and the decision for accepting or rejecting the treatment model based on them. It is evident from the results that Macro and Micro methods changed their decision with very drastic drop in their statistics after the traffic was increased to the treatment model. The proposed method however, suggests to reject the model in both experiments. Essentially, in both experiments we observe that the Z-test accepts the null hypothesis that the model has no effect. Even though in the second experiment the confidence interval is more towards the negative effect, it is not significant given the required significance level. Even if we declare the summary effect significant by reducing the significance level, the confidence interval still suggests a reject.

Figure 4 illustrates the difference of the ROIs of the control and treatment models for the two traffic settings for each campaign versus its *total* spend. Each dot represents a campaign. The blue dots stand for 10% and red ones for 20% traffic designation to the treatment model. We have removed the numbers in the x-axis to comply with company's policies but obviously the right most dots represent the highest spending campaigns. Since we have pruned the campaigns for only those that are common between the two experiments and have no setup change, each blue dot has a corresponding red dot with almost the same x coordinate (spend). Note that many red dots are positioned below the blue ones. Some of them are in the right half of the plot, the highest spending campaigns. This explains why Macro and Micro changed their decisions after the traffic was increase. Micro considers a weighted average of the differences where weights are determined by the spend of each campaign. This makes this method particularly sensitive to the behavior of high spending campaigns (the right half) and thus its statistics dropped substantially after the traffic increase.

TABLE I. RESULTS OF APPLYING EACH METHOD ON COMMON STABLE CAMPAIGNS

| | Parameter | Statistics | | Decision | |
|---|---|---|---|---|---|
| Treatment traffic | | 10% | 20% | 10% | 20% |
| Macro | $\theta = 0.004$ | $\mu^{Macro} = 0.17$ | $\mu^{Macro} = -0.05$ | Accept | Reject |
| Micro | $\theta = 0.01$ | $\mu^{Micro} = 0.29$ | $\mu^{Micro} = -0.31$ | Accept | Reject |
| Proposed Method | $\alpha = 95\%$ | $CI = [-0.02, 0.01]$ $P_Z = 25.7\%$ | $CI = [-0.028, 0.009]$ $P_Z = 15.7\%$ | Reject | Reject |

**Setting 2:** In the second set of experiments we apply the methods on all campaigns regardless of their presence in both experimentation time periods or their setup. The results are reported in Tab. II. Again, we observe the same trend. All competing methods report substantial drop in the measure and change in their decision while the proposed method suggests to reject the model in both cases due to ineffectiveness. As in the previous setting, we did not observe any change in the parameters based on the A/A test.

The success of the proposed approach lies in the way that it chooses to weigh the effect of each campaign and the fact that it considers the variations within and between the studies. It is important to note that we need to consider the outliers in our final decision making since they are still part of our system. The fact that they have served enough impressions while the treatment model had an extraordinary effect on them makes them interesting but does not allow us to ignore them. Smaller traffic increases our uncertainty about the observed effect and also a large effect must be downgraded as it can be easily caused by sampling error. All of these intuitions are captured in the weighting mechanism of the proposed method.

To observe why Macro-averaging changes its decision given the very same parameters, refer to Fig. 5. The reported histograms in Fig. 5 shows the distribution of the difference of the ROIs for all campaigns. Note that when the traffic directed to the treatment is at 10% (left plot), the histogram is tilted towards positive differences of ROI with a very prominent outlier. Once the traffic is increased to 20% (right plot), the distribution becomes narrower (smaller variance) and slightly tilts towards the negative differences. This results into a change of direction in the decision of the Macro methods.

Micro-averaging, as discussed in the previous setting, is susceptible to outliers when they happen to have high budgets. This property of Micro-averaging, prevents it to be fair to many campaigns that have medium or lower budgets but it is also desirable in the sense that it brings some notion of business value into the analysis. In other words, the high spending clients tend to be more important from a business perspective. To add an analogy to our method, we resort to subgroup analysis, which studies the behavior of the treatment model in more depth.

We need to mention that the homogeneity test for all experiments failed to reject the $\chi^2$ test with 10% significance level and thus, the effect was considered to be homogeneous overall (loosely speaking).

To study the importance of the spend (budget) on the variation of the ROI effect sizes, we proceed as follows: We divide the campaigns into three subgroups based on their spend. We sort the campaigns decreasingly by their spend and use the cumulative sum of the spend to divide them into subgroups. Subgroup 1 contains the highest spending campaigns that collectively spend 33% of the total amount of the money spent by both models and subgroup 3 is comprised of the lowest spending ones. Subgroup 3 contains much more campaigns compared to the other two. The results are shown in Tab. III for both 10% and 20% traffic shares and all running campaigns. The subgroup analysis reveals that regardless of the budget, all campaigns suffer some degradation in the performance where $P_{Q_{between}} = 8\%$ is under 10% traffic and it decreases to 4% in case of 20% traffic, which is significant for a 10% significance level. This means that the group membership affects the variation of the effect size and the effect of the model on each campaign is dependent on the given budget. Note that subgroup 2 is more severely affected by the new model. Basically, from the subgroup analysis based on the spend we conclude that there is no benefit in applying the model to a certain group of campaigns and it is particularly harmful to medium budget campaigns.

In summary, the proposed approach enabled us to reject an ineffective model in earlier stages compared to other methods and provided us with various insights into the behavior of the model in a concise set of statistics. We observed how the treatment model affected the overall performance of the system through robust and quantitative measurements. Furthermore, we studied its behavior given a more similar set of campaigns through subgroup analysis and observed that its insignificant performance can only get worse for medium budgeted campaigns and there is no chance of obtaining any improvement for any categories of spend.

## VII. CONCLUSION

In this paper we presented a general framework that could reliably and consistently evaluate a new model compared to a baseline for online bid prediction in a very large system comprised of diverse set of campaigns with variety of goals and budgets. We described the challenges in this area and variety of practices and ideas that we had implemented to arrive at the current system. Our system is able to efficiently and dynamically direct the online traffic to appropriate experimental and baseline models allowing us to have a statistically sound randomization and thus evaluation.Using the return of investment (ROI) as our unified performance metric and the implemented system, we showed, using an example, that our proposed framework is able to recognize the quality of a new model in very early stages leading to a proper decision for rejecting it while other method needed much more data to arrive at the same conclusion.

TABLE II. RESULTS OF APPLYING EACH METHOD ON ALL CAMPAIGNS

| | Parameter | Statistics | | Decision | |
|---|---|---|---|---|---|
| Treatment traffic | | 10% | 20% | 10% | 20% |
| Macro | $\theta = 0.001$ | $\mu^{Macro} = 0.14$ | $\mu^{Macro} = -0.03$ | Accept | Reject |
| Micro | $\theta = 0.005$ | $\mu^{Micro} = 0.10$ | $\mu^{Micro} = -0.99$ | Accept | Reject |
| Proposed Method | $\alpha = 95\%$ | $CI = [-0.019, 0.009]$ $P_Z = 24.2\%$ | $CI = [-0.03, 9.6 \times 10^{-4}]$ $P_Z = 3.3\%$ | Reject | Reject |

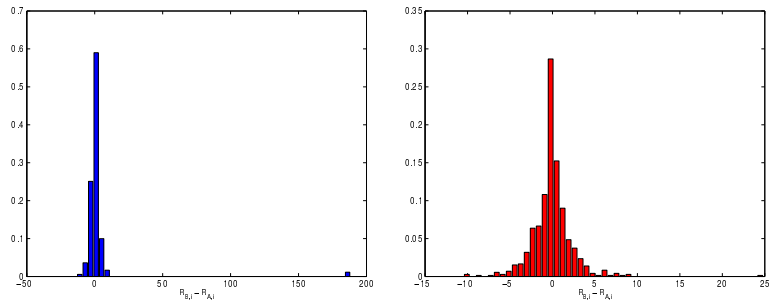

Fig. 5. The histogram of the difference of ROIs reported by model A and B for *all* campaigns under 10% (left) and 20% (right) traffic experiments. The Macro method changes it decision because the distribution of the difference of ROIs slightly tilts towards more negative values when traffic share increases to 20%.

TABLE III. STATISTICS FOR THE SUBGROUP ANALYSIS OF ROI

| | Subgroup 1 | | Subgroup 2 | | Subgroup 3 | |
|---|---|---|---|---|---|---|
| Traffic share | 10% | 20% | 10% | 20% | 10% | 20% |
| $CI$ | $[-0.26, 0.19]$ | $[-0.21, 0.11]$ | $[-0.12, 0.05]$ | $[-0.17, 0.02]$ | $[-0.02, 0.02]$ | $[-0.02, 0.01]$ |
| $P_Z$ | 38% | 27% | 21% | 6.1% | 50% | 25.7% |
| $p_{Q^*}$ | 43% | 36% | 41% | 9.1% | 35% | 99% |